\documentclass[10.5pt,compsoc]{BD}
\usepackage{graphicx}
\usepackage{footmisc}
\usepackage{subfigure}
\usepackage{url}
\usepackage{multirow}
\usepackage[noadjust]{cite}
\usepackage{amsmath,amsthm}
\usepackage{amssymb,amsfonts}
\usepackage{booktabs}
\usepackage{color}
\usepackage{ccaption}
\usepackage{float}
\usepackage{fancyhdr}
\usepackage{caption}
\usepackage{xcolor,stfloats}
\usepackage{cuted}
\usepackage{captionhack}
\usepackage{epstopdf}
\usepackage{xcolor,colortbl}

\headevenname{\zihao{-5}{\textbf{\emph{Big Data Mining and Analytics, January}}} 2018, 1(1): 000-000}%
\headoddname{{\sf Zu et al.:}\quad {\textbf{\emph{Memory-Efficient Gradient Computation}}}}%

\newtheoremstyle{mystyle}{0pt}{0pt}{\normalfont}{1em}{\bf}{}{1em}{}
\theoremstyle{mystyle}
\renewcommand\figurename{Fig.~}

\begin{document}

%%%%%%%%%%%%%%%%%%%%%%%%%%%%%%%%%%%%%%%%%%%%%%%%%%%%%%%%%%%%%%
% TITLE BLOCK
%%%%%%%%%%%%%%%%%%%%%%%%%%%%%%%%%%%%%%%%%%%%%%%%%%%%%%%%%%%%%%
\begin{strip}\zihao{3}
\centering
% Memory-Efficient Gradient Optimization for On-device Training
% Toward On-device Training: Benchmarking Memory-Efficient Gradient Computation Methods.
% Toward On-device Training: Benchmarking Memory-Efficient Gradient Computation
% XXXX Gradient under the Microscope: Benchmarking Memory-Efficient Gradient Computation
% Optimizing Under Constraint: A Comparative Study of Gradient-Efficient Fine Tuning
\textbf{ Gradient Under Microscope: Benchmarking Resource Utilization of
Memory-Efficient Gradient Computation Methods            }
\vskip 6mm
\zihao{5}

\centering {\sf 
Sarthak Mahapatra*, Zihan Zhou*, Khatoon Khedri, Mehdi Hosseinzadeh, Reza Rawassizadeh}
\vskip 5mm

% possible workshop: https://wdlctc.github.io/efficient-reasoning-2026/#introduction

% Abstract
\begin{tabular}{p{160mm}}
{\zihao{-5}
\noindent
\bf{Abstract:} {\sf 
A major bottleneck of AI model adaptation is their resource utilization. Large AI models run in data centers with a vast carbon footprint.  
% Besides, due to their complexity they need to run on the resource-full cloud system and deliver the output to their users. Delivering data to the cloud of a large corporation is associated with privacy risks. Local training of AI models offers privacy benefits and less network overhead, but it faces memory and computational constraints.
We evaluate gradient optimization strategies for efficient training on resource-limited hardware. We compare five gradient optimizers (Stochastic and Conjugate Gradient Descent, Adam, Adagrad, and Adadelta) under three memory configuration strategies, baseline training, gradient checkpointing, and gradient accumulation, across four different architectures (ViT, ModernBERT, Llama\,3.1\,1B, and NanoVLM). 

We report training loss, GPU utilization, training time, and memory usage. Our results show that gradient accumulation is the most reliable technique, reducing training loss by roughly an order of magnitude on the VLM and about 4$\times$ on the LLM without additional GPU memory. Contrary to common practice, the adaptive optimizer Adam is not universally superior, it is outperformed by Adadelta and SGD on the encoder (ModernBERT) and autoregressive (LLM) models. Gradient checkpointing is strongly architecture-dependent, it lowers ViT loss but degrades ModernBERT severely, while increasing training time substantially (by up to 60\% on the memory-bound models). GPU utilization is governed primarily by architecture, ranging from 8--15\% for the memory-bound LLM to 96--99\% for the compute-bound vision models. Results of our experiments provide practical guidelines for optimizer selection and  resource-efficient model training and deployment.
}
}
\end{tabular}
\end{strip}

\section{Introduction}
Most countries including United States \cite{muller2026measuring} are facing electricity shortages. At the same time, there is massive investment in AI data centers that consume a huge amount of electricity. Training deep neural network architectures has become increasingly resource-intensive as models grow in size, complexity, and deployment scope. Large Language Models (LLMs), Vision Transformers (ViTs), and even lightweight, yet high-performance architectures, such as ModernBERT \cite{modernbert} and NanoVLM \cite{nanovlm} now form the backbone of natural language processing, computer vision, and multimodal reasoning systems \cite{vaswani2017attention, devlin2019bert, dosovitskiy2020vit}. As these models scale, their computational and memory requirements escalate correspondingly, creating significant challenges for researchers and practitioners operating under constrained hardware environments. 
Consequently, understanding how training techniques and optimization strategies influence memory usage, computational load, and convergence behavior is a critical practical concern for democratizing AI development beyond well-funded corporations.

A substantial body of prior work highlights the importance of improving training efficiency in large-scale neural networks. The foundational transformer architecture \cite{vaswani2017attention} is known for the high memory cost of self-attention, motivating subsequent research into efficient optimization and memory-aware training. Optimizer design has played a central role in this effort. Adaptive methods, such as Adam \cite{kingma2014adam} and Adadelta \cite{zeiler2012adadelta}, are known to accelerate convergence, while simpler non-adaptive methods, like Stochastic Gradient Descent (SGD), remain attractive due to their lower memory footprint and predictable behavior. However, the trade-offs between convergence speed, stability, and memory usage vary significantly across architectures and training regimes.

There are efforts that explore gradient-level techniques to mitigate memory and compute bottlenecks. Gradient checkpointing \cite{checkpointing} reduces activation memory by recomputing intermediate states during backpropagation, while gradient accumulation enables larger effective batch sizes without increasing peak memory usage \cite{ott2018scaling}. Additionally, second-order or quasi-second-order methods, including conjugate gradient-based approaches \cite{martens2010deep}, have been investigated for their theoretical convergence properties, though their empirical behavior in large transformer models remains less well understood. Despite these advances, systematic empirical comparisons of such techniques across fundamentally different model families, such as LLMs, encoder-based language models, vision transformers, and multimodal architectures, are still limited.

The diversity in model architectures further complicates the choice of optimization strategy. Decoder-only LLMs exhibit large activation footprints and long gradient paths, making them especially sensitive to memory-efficient techniques \cite{shoeybi2019megatron}. Vision Transformers rely heavily on attention-based feature aggregation, which introduces different computational and memory bottlenecks compared to convolutional models \cite{dosovitskiy2020vit}. Lightweight architectures such as ModernBERT \cite{modernbert} and NanoVLM \cite{nanovlm} are designed for efficiency, yet their interaction with adaptive optimizers and gradient-level techniques has not been extensively characterized in a unified experimental framework. These architectural differences motivate a comprehensive empirical study that evaluates optimization behavior across domains rather than in isolation.

Detailed carbon footprint data is available for large, openly-documented models like Llama 2 (539 tCO2eq for 3.3M GPU hours) \cite{llama2}, Llama 3 (11,390 tCO2eq) \cite{llama3}, and Llama 4 (1,999 tCO2eq for training) \cite{llama4}. Also, detailed carbon footprint data is available for fine-tuning experiments on architectures such as Google ViT (0.066 kg CO2eq per experiment) \cite{ordoumpozanis2024green}. However, these benchmarks are model specific and to our knowledge there is no benchmark analysis of different gradient-efficient computation methods. This issue underscores the need for the kind of systematic efficiency analysis, considering memory, time, and hardware utilization.

Furthermore, the practical deployment of even middle size AI models on edge devices such as smartphones or mobile robots are fundamentally constrained by three interlinked resources: computation budget, memory capacity, and battery power and thermal constraints \cite{odsearch, xu2024ondevice, zheng2025edgellm}. A model's forward pass, required for inference, demands storing large weight matrices and intermediate activations—the very activations whose gradient computation we study during training \cite{neuroviz}. This footprint can easily exceed an edge device's available memory. For instance, even a 3B parameter model can require ~8 GB of memory, while a smartphone device typically have only 5–9 GB of usable RAM after operating system overhead \cite{liu2025m}. Furthermore, memory bandwidth on mobile devices (50–90 GB/s) is 20–50× lower than on data center GPUs (2–3 TB/s), making data movement the primary bottleneck rather than compute \cite{One-device-LLM}. These hardware limitations necessitate the memory-efficient techniques investigated in this work.

We pair these models with five widely used optimizers (baseline SGD \cite{sgd}, Adam \cite{kingma2014adam}, Adadelta \cite{zeiler2012adadelta}, and Adagrad \cite{adagrad}), together with Conjugate Gradient Descent (CGD) \cite{cgd}, a resource-efficient gradient computation approach \cite{rezabook}. For the NanoVLM experiments, we additionally evaluate AdamW \cite{loshchilov2019adamw}, which decouples weight decay from the adaptive optimization step and has become the standard optimizer for transformer-based vision-language models. We further evaluate three gradient optimization strategies: baseline gradient descent, gradient checkpointing \cite{checkpointing}, and gradient accumulation \cite{accumulation}, which are used to reduce the memory cost of gradient descent \cite{rezabook}.

Together, this design yields a single experimental framework in which memory-efficient methods, adaptive optimizers, and theoretically motivated alternatives such as CGD are evaluated under identical conditions. Beyond model loss and convergence behavior, we systematically monitor system-level metrics, including GPU utilization, memory (RAM) usage, and training time per epoch, across all four architectures. Prior studies have emphasized that such hardware-centric measurements often determine the feasibility of training and deploying large models in edge and resource-constrained environments \cite{lane2016deepx, han2016deep}. By jointly analyzing convergence and hardware efficiency, we produce the cross-architecture efficiency benchmark that is currently missing from the literature.

Overall, our work bridges the gap between theoretical optimization techniques and their empirical behavior in modern transformer-based models. Our findings offer actionable guidance for selecting optimizer and gradient strategies that balance performance, memory efficiency, and computational cost, enabling more informed training decisions under real-world hardware constraints. We hope our study facilitates the democratization of AI development, enabling small organizations and independent researchers to build state-of-the-art AI models.

\section{Methods}
To investigate how different gradient computation strategies interact with model architecture, we design a controlled experimental framework spanning four representative transformer-based architectures, five optimizers, and three memory-efficient training techniques. This section details the models, optimizers, and training strategies evaluated in our study, following recent calls for cross-architecture empirical analysis under resource-constrained conditions.

\subsection{Models}
We evaluate four transformer-based architectures that collectively represent language, vision, and multimodal learning, emphasizing cross-architecture evaluation under resource constraints \cite{shallue2018measuring}.

\textbf{Llama\,3.1\ (1B)} is a compact (1 billion parameter) autoregressive language model used to study large-scale decoder-only transformer training under limited GPU memory. Decoder-only architectures exhibit large activation footprints and long gradient paths, making them highly sensitive to optimizer choice and memory-efficient training strategies \cite{dettmers2022efficient, rajbhandari2020zero}.

\textbf{ModernBERT} is a lightweight encoder-only transformer reflecting recent architectural trends toward efficiency-oriented NLP models. Encoder-based transformers are generally more stable during training and exhibit lower memory sensitivity, providing a useful contrast to large autoregressive models \cite{devlin2019bert}.

\textbf{NanoVLM (Vision--Language Model)} is a compact multimodal architecture that integrates visual embeddings with textual representations via cross-attention. Multimodal models introduce heterogeneous activation patterns and increased memory pressure, making them well suited for evaluating optimization and memory trade-offs in constrained environments \cite{ivanov2022benchmarking}.

\textbf{Google ViT} is a vision transformer that processes images as patch embeddings and applies standard transformer blocks. Vision transformers are known to shift computational bottlenecks toward attention operations and benefit differently from optimization strategies compared to language models \cite{narang2021transformer}.

Together, these four models span decoder-only LLMs, encoder-based language models, multimodal architectures, and vision transformers, enabling a comprehensive evaluation of training efficiency across domains.

\subsection{Optimizers}

We evaluate five widely used optimizers that reflect practical trade-offs between convergence speed, stability, and memory consumption, as highlighted in prior empirical and systems-level studies \cite{shallue2018measuring}.

\textbf{Stochastic Gradient Descent(SGD)} serves as a fundamental baseline due to its minimal memory footprint and predictable optimization behavior. Despite slower convergence, SGD remains attractive in memory-constrained environments and large-batch training regimes \cite{you2017scaling, smith2018disciplined, shallue2018measuring}.

\textbf{Adam} \cite{kingma2014adam, dettmers2022efficient} is a widely adopted adaptive optimizer that leverages momentum and per-parameter learning rates to accelerate convergence in non-convex optimization landscapes. Its effectiveness comes at the cost of increased memory usage due to additional moment estimates, which can limit scalability under tight hardware constraints .

\textbf{Adagrad} \cite{adagrad} adapts learning rates based on accumulated squared gradients and is particularly effective for sparse or noisy gradients commonly observed in NLP workloads. However, its growing memory requirements over training time motivate careful evaluation under constrained settings.

\textbf{Adadelta} \cite{zeiler2012adadelta} addresses Adagrad aggressive learning-rate decay issue, by using a moving window of gradient statistics, offering more stable long-term optimization behavior. It represents a middle ground between fully adaptive methods and simpler first-order optimizers \cite{shallue2018measuring}.

\subsection{Memory Efficient Techniques}
Following prior systems and optimization-focused research \cite{ivanov2022benchmarking, checkpointing, rezabook}, we evaluate three training strategies aimed at reducing memory overhead and improving training feasibility under limited hardware resources.

\textbf{Gradient Checkpointing} reduces activation memory by storing only a subset of intermediate activations during the forward pass and recomputing others during backpropagation. While effective at lowering peak memory usage, this technique introduces additional computational overhead and altered GPU utilization patterns\cite{checkpointing}.

\textbf{Gradient Accumulation} increases the effective batch size by accumulating gradients across multiple iterations before applying an optimizer update. This technique improves gradient stability and convergence behavior without increasing peak memory usage, making it particularly suitable for large models trained on limited hardware \cite{shallue2018measuring, smith2018disciplined}.

\textbf{Conjugate Gradient Descent (CGD)} is a second-order-inspired optimization method that updates parameters along conjugate directions, potentially improving convergence in well-conditioned settings. However, its computational complexity and sensitivity to model conditioning make its practical behavior in transformer training an open empirical question \cite{martens2010deep}.

\section{Experimental Setup}

\begin{figure*}[!t]
\centering
\includegraphics[width=\linewidth]{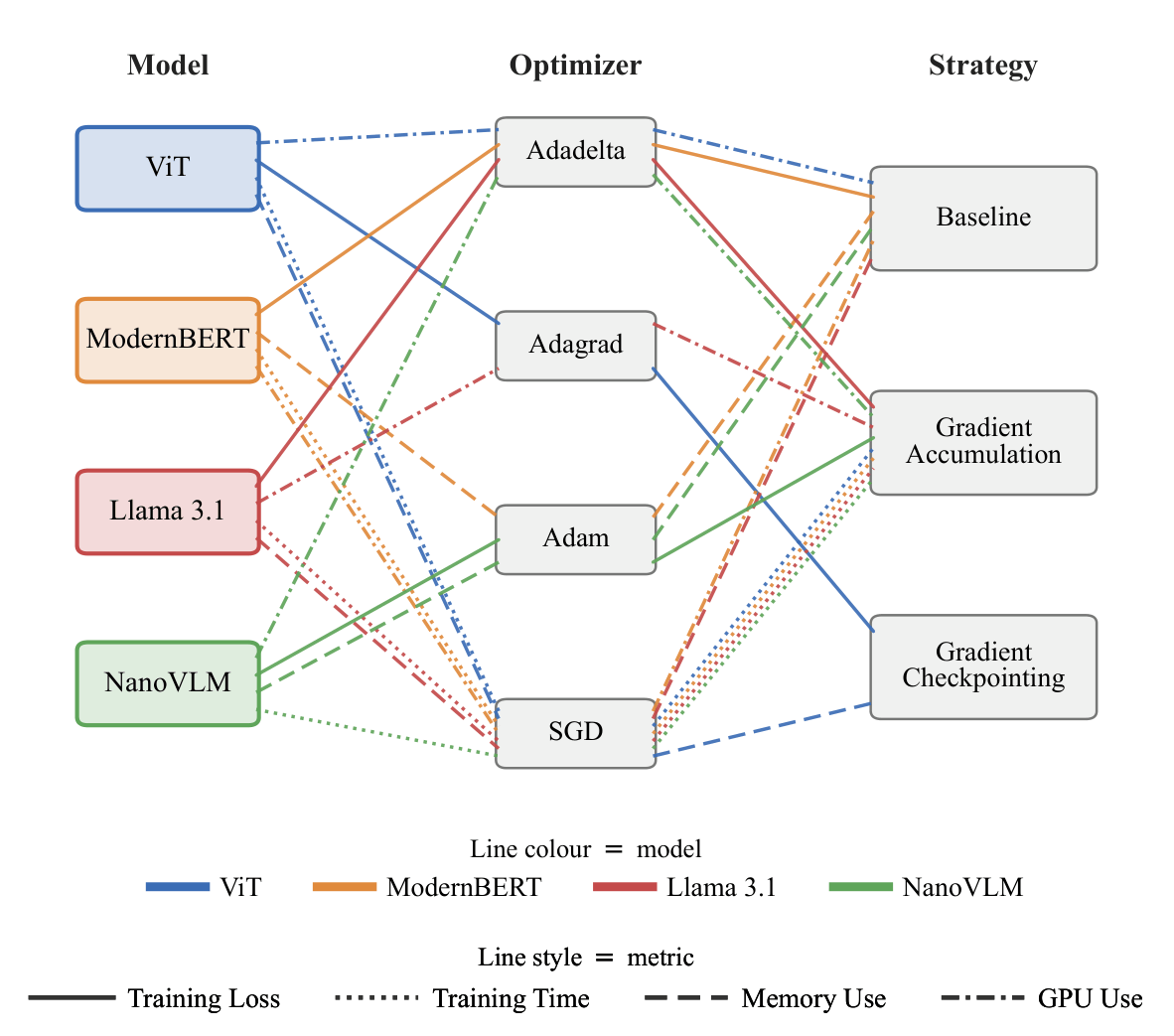}
\caption{Overview of the experimental design space and headline result. The three columns enumerate the choices evaluated in our study, including model architectures, optimizers, and gradient-memory strategies. Each colored line traces the empirically \emph{best} optimizer strategy configuration for one (model, metric) pair. Conjugate Gradient Descent (CGD) was also evaluated but is never the best on any pair, so it is omitted from the optimizer column.}
\label{fig:best-combos}
\end{figure*}

Figure~\ref{fig:best-combos} provides an overview of the experimental pipeline, including model selection, optimizer and gradient strategy choices, and the metrics monitored during training. 

All experiments were conducted on a dedicated Linux workstation equipped with a single NVIDIA A100 GPU (40GB VRAM) and 256GB system RAM. This hardware configuration supports full-scale fine-tuning of all selected architectures while still reflecting the constraints typically encountered in academic environments. 

ViT, ModernBERT, and Llama\,3.1\,1B were each trained for 25 epochs under every optimizer and gradient memory configuration. NanoVLM, due to its larger per-step cost, was trained for a single epoch (18{,}623 training steps), with metrics logged at every step to provide fine-grained convergence profiles. To enable cross-model comparison, we report NanoVLM results aggregated over 25 equally spaced bins, each corresponding to a pseudo-epoch of approximately 745 steps. Within each model, training duration and computational budget were held constant across all optimizer technique combinations, ensuring the fair setting and observed differences arise from optimization behavior alone.

Table~\ref{hw_budget} summarizes the theoretical compute budget of the A100 platform across the numerical precisions relevant to deep-learning training. These numbers represent the data-center capacity against which edge hardware must be contrasted, mobile accelerators typically deliver a few to tens of INT8 TOPS with 5-9\,GB of shared memory and 50-90\,GB/s of bandwidth \cite{liu2025m, One-device-LLM}, one to two orders of magnitude below the A100. 

\begin{table}[htbp]
\centering
\caption{NVIDIA A100 (40\,GB) Compute Budget. Tensor Core throughput is reported at peak; values in parentheses use structural sparsity.}
\label{hw_budget}
\small
\setlength{\tabcolsep}{5pt}
\begin{tabular}{lc}
\toprule
\textbf{Resource} & \textbf{Capacity} \\
\midrule
FP32                       & 19.5 TFLOPS \\
TF32 Tensor Core           & 156 (312) TFLOPS \\
FP16/BF16 Tensor Core      & 312 (624) TFLOPS \\
INT8 Tensor Core           & 624 (1248) TOPS \\
GPU memory (HBM2e)         & 40 GB \\
Memory bandwidth           & 1{,}555 GB/s \\
Thermal design power       & 400 W \\
\bottomrule
\end{tabular}
\end{table}

We used a standardized dataset \cite{pranavmr2025mmimdb} across model families, consisting of a text corpus for the language models (Llama\,3.1\,1B. and ModernBERT) and an image-text paired dataset for the vision models (NanoVLM and ViT). The text data were tokenized into fixed-length sequences of 512 tokens, while all images were preprocessed to $224\times224$ pixels. For NanoVLM, captions were truncated or padded to 128 tokens. All datasets were shuffled, batched uniformly, and split consistently across runs to eliminate variation due to data ordering.

Each model was fine-tuned using a stable set of hyperparameters. Llama\,3.1\,1B was trained with a learning rate of $2 \times 10^{-5}$, batch size 16 (or effective batch size 64 with accumulation), a warmup ratio of 0.03, and a maximum sequence length of 512. ModernBERT uses a learning rate of $1 \times 10^{-4}$, batch size 32 (or effective batch size 128 with accumulation), GELU activations, and LayerNorm applied to all hidden states. NanoVLM was trained with a learning rate of $5 \times 10^{-5}$, batch size 32, and cross-entropy loss over caption tokens. ViT-Base, configured with a patch size of 16, hidden dimension 768, and 12 attention heads, was fine-tuned using a learning rate of $3 \times 10^{-4}$ and batch size 64 (or effective batch size 256 when accumulation was enabled). No hyperparameters were altered between optimizer-technique combinations, ensuring a controlled and isolated comparison of methods.

For each model, we evaluated a common set of optimizers including Adam, Adadelta, SGD, and Adagrad combined with three gradient memory strategies: standard training, gradient accumulation, and activation checkpointing. In addition, Conjugate Gradient Descent (CGD) was included for ViT, ModernBERT, and Llama\,3.1\,1B, where its convergence behavior could be assessed over a full 25-epoch schedule. For NanoVLM, we additionally evaluated AdamW alongside Adam to examine the effect of decoupled weight decay in a vision-language setting, as AdamW separates weight decay from the adaptive optimization step and has been shown to improve optimization and generalization compared with conventional Adam.

During training, we continuously collected both system resources and model-level metrics. System resources included GPU memory usage, GPU utilization, host RAM consumption, and wall-clock training time. Model metrics included training loss and, where applicable, accuracy. For epoch-based models (ViT, ModernBERT, Llama\,3.1\,1B), metrics were recorded once per epoch; for NanoVLM, metrics were recorded at every training step. These measurements were gathered across all configurations, enabling comparison not only of convergence behavior but also of computational overhead and memory efficiency.

By holding hyperparameters, dataset splits, and training budget constant within each model, the experiments isolate the true performance, stability, and resource trade-offs introduced by checkpointing, gradient accumulation, and CGD. All summary tables report metrics averaged over the full training run (25 epochs or 18{,}623 steps, as applicable). The resulting observations highlight which combinations of optimizers and techniques are most suitable for scenarios where memory availability, computational speed, or convergence stability are the primary constraints.

\section{Results}
As described before, we have evaluated three gradient computation strategies, including baseline Backpropagation, Gradient Checkpointing, and Gradient Accumulation, across four diverse architectures (ViT, ModernBERT, LLM, and VLM) using up to five optimizers (SGD, Adam, Adadelta, Adagrad, and CGD where applicable; AdamW additionally for VLM). Our experiments reveal interesting patterns in how these strategies interact with different model families.

\subsection{Training Loss}

\begin{figure*}[!t]
\centering
\includegraphics[width=\linewidth]{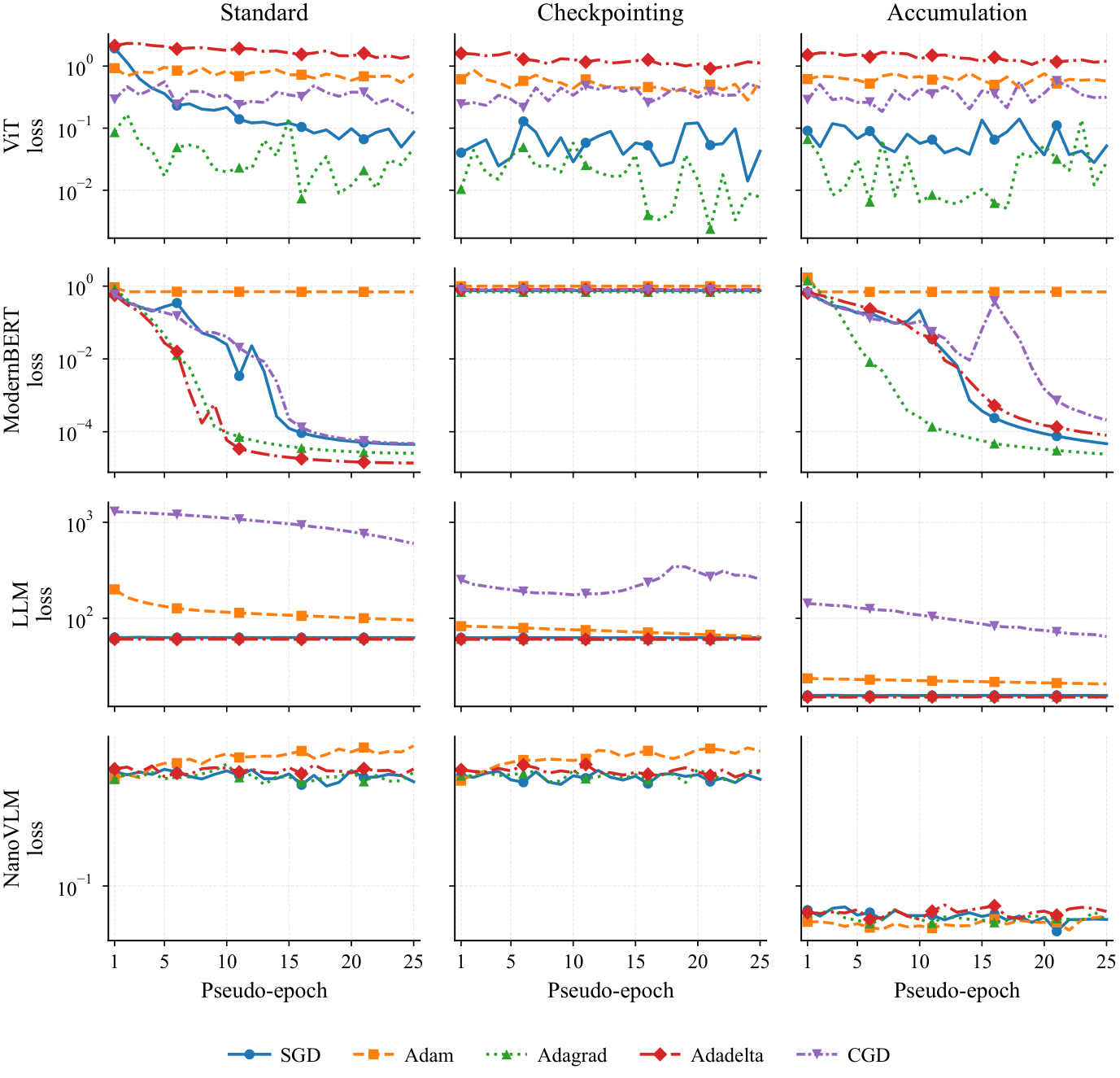}
\caption{Training loss trajectories across the four model architectures (rows) and three gradient strategies (columns). All panels use a logarithmic y-axis; y-limits are shared within each row to support cross-strategy comparison.}
\label{fig:loss-curves}
\end{figure*}

The most consistent finding across all architectures is the substantial improvement in training loss achieved by gradient accumulation.NanoVLM step-level data has been aggregated into 25 pseudo-epochs to match the training schedule of the other models. Gradient accumulation lowers the loss floor for ViT, LLM, and NanoVLM; checkpointing has a strong regularizing effect on ViT but severely degrades ModernBERT. Figure~\ref{fig:loss-curves} shows the per-epoch loss trajectories for every (model, optimizer, strategy) combination. Table~\ref{best_loss} summarizes the best-performing configurations for each model. We investigate the impact of all gradient computations on the listed models and describe them as follows.

\begin{table}[htbp]
\centering
\caption{Best Training Loss Achieved by Each Model}
\label{best_loss}
\small
\setlength{\tabcolsep}{4pt}
\begin{tabular}{lcccc}
\toprule
\textbf{Model} & \textbf{Best Loss} & \textbf{Optimizer} & \textbf{Strategy} & \textbf{Improvement} \\
\midrule
ViT          & 0.02   & Adagrad   & Checkpointing & 2.0$\times$ \\
ModernBERT   & 0.05   & Adadelta  & Standard      & --- \\
LLM          & 15.14  & Adadelta  & Accumulation  & 4.0$\times$ \\
VLM          & 0.06   & Adam      & Accumulation  & 11$\times$ \\
\bottomrule
\end{tabular}
\end{table}
\textbf{Gradient Accumulation.} For VLM, accumulation reduces average loss by roughly an order of magnitude (e.g., Adam drops from 0.66 to 0.06, and Adagrad from 0.50 to 0.06). For the LLM, the improvement is similarly pronounced, with loss dropping from 60+ to around 15. ViT exhibits more moderate gains; SGD improves from 0.28 to 0.07, while Adagrad's already low standard loss (0.04) leaves little room for further reduction. The exception is ModernBERT, where standard training already achieves excellent loss (0.05), and accumulation performs slightly worse (0.12). This broad benefit suggests that gradient variance is a fundamental challenge for transformer-based architectures. The self-attention mechanism, which computes pairwise interactions between all tokens, may amplify gradient noise when batch sizes are small. Accumulation effectively simulates larger batches, producing more stable gradient estimates.

\textbf{Gradient Checkpointing.} Unlike accumulation, gradient checkpointing shows strongly architecture-dependent effects. Table \ref{checkpointing_impact} compares checkpointing performance relative to standard training.
\begin{table}[htbp]
\centering
\caption{Impact of Gradient Checkpointing (Loss Ratio: Checkpointing / Standard). Bold values indicate that checkpointing reduces loss ($<1.0$); values $>1.0$ indicate degradation.}
\label{checkpointing_impact}
\small
\setlength{\tabcolsep}{5pt}
\begin{tabular}{lcccc}
\toprule
\textbf{Optimizer} & \textbf{ViT} & \textbf{ModernBERT} & \textbf{LLM} & \textbf{VLM} \\
\midrule
Adadelta & \textbf{0.69} & 16.0  & 1.00          & 1.00 \\
Adagrad  & \textbf{0.50} & 9.9   & 0.98          & 1.00 \\
Adam     & \textbf{0.68} & 1.41  & \textbf{0.62} & 1.00 \\
CGD      & 1.06          & 9.9   & \textbf{0.24} & N/A \\
SGD      & \textbf{0.21} & 7.6   & 1.00          & 1.00 \\
\bottomrule
\end{tabular}
\end{table}
For ViT, checkpointing consistently improves or maintains loss across all optimizers, with SGD showing the largest benefit (0.28$\rightarrow$0.06, a 79% reduction). For the VLM, checkpointing has no measurable effect on average loss, with ratios near 1.00 across optimizers. These results suggest that activation recomputation introduces minimal numerical disruption in vision and vision-language models and may even provide a modest regularization benefit in ViT.

In contrast, checkpointing severely degrades ModernBERT, increasing loss by 7.6--16$\times$ across all optimizers. Adadelta, for example, achieves a loss of 0.05 with standard training but rises to 0.80 with checkpointing. This degradation may result from ModernBERT's deep computation graph and the sensitivity of its layer-normalization and attention operations to small numerical differences introduced during activation recomputation. Therefore, checkpointing should be used cautiously with encoder-only architectures, and its effect should be empirically validated before deployment.

For the LLM, checkpointing shows optimizer-dependent effects. Loss improves from 118.22 to 73.75 (38\% reduction) for Adam, a clear benefit. For Adadelta, Adagrad, and SGD, losses are nearly identical with checkpointing ($\pm$2\%). CGD improves substantially but remains unusable, dropping from 996.59 to 234.31, still 3-4$\times$ worse than other optimizers with checkpointing. The improvement for Adam suggests that its adaptive mechanism may be more robust to recomputation, while simpler methods treat the recomputed gradients as sufficiently accurate.

\textbf{Standard Backpropagation.} Our results reveal several surprises about standard optimizer performance that challenge common practices in deep learning.
Adam is Not Always Optimal. Despite being the default choice for most transformer applications, Adam underperforms across multiple architectures. In ModernBERT, Adam (0.71) vs.\ Adadelta (0.05)---Adam is 14$\times$ worse. For the LLM, Adam (118.22) vs.\ Adadelta (60.54)---Adam is 2$\times$ worse. In ViT, Adam standard (0.75) vs.\ Adagrad standard (0.04)---Adam is nearly 19$\times$ worse.
This challenges the assumption that adaptive methods are universally superior. For encoder-only models such as ModernBERT, the simpler optimization landscape may favor methods with less aggressive adaptation.
Adagrad achieves the lowest loss on ViT across all strategies, reaching 0.02 with checkpointing---the best of any ViT configuration. Adagrad's accumulation of squared gradients in its denominator benefits from stable gradient estimates, providing both adaptive learning rates and low-variance gradients.

Adadelta achieves the best or near-best loss on ModernBERT (0.05, standard) and LLM (15.14, accumulation). For VLM, several optimizers achieve comparable accumulation losses near 0.06--0.07, with Adam marginally lowest (0.06). This consistency suggests that Adadelta's moving-window approach to gradient history provides a good balance of adaptation and stability across diverse tasks.

CGD shows severe failure on the LLM, with losses of 996 (standard), 234 (checkpointing), and 99 (accumulation). Even with accumulation, CGD's loss is 5–6X higher than other optimizers. This indicates that second-order methods, in their raw form, are unsuitable for autoregressive language modeling. The curvature information is either too expensive to compute accurately or too unstable to be useful.

% ------------------------ VIT LOSS ------------------------

\begin{table}[H]
\centering
\caption{ViT Training Loss Across Gradient Strategies}
\label{Loss-VIT}
\begin{tabular}{lccc}
\toprule
\textbf{Optimizer} & \textbf{Standard} & \textbf{Checkpointing} & \textbf{Accumulation} \\
\midrule
Adadelta & 1.78 & 1.23 & 1.38 \\
Adagrad  & 0.04 & \cellcolor{green!25}0.02 & 0.03 \\
Adam     & 0.75 & 0.51 & 0.62 \\
CGD      & 0.34 & 0.36 & 0.35 \\
SGD      & 0.28 & 0.06 & 0.07 \\
\bottomrule
\end{tabular}
\end{table}

% ------------------------ MODERNBERT LOSS ------------------------

\begin{table}[H]
\centering
\caption{ModernBERT Training Loss Across Gradient Strategies}
\label{Loss-ModernBERT}
\begin{tabular}{lccc}
\toprule
\textbf{Optimizer} & \textbf{Standard} & \textbf{Checkpointing} & \textbf{Accumulation} \\
\midrule
Adadelta & \cellcolor{green!25}0.05 & 0.80 & 0.12 \\
Adagrad  & 0.07 & 0.69 & 0.10 \\
Adam     & 0.71 & 1.00 & 0.74 \\
CGD      & 0.08 & 0.79 & 0.12 \\
SGD      & 0.10 & 0.76 & 0.11 \\
\bottomrule
\end{tabular}
\end{table}

% ------------------------ LLM LOSS ------------------------

\begin{table}[H]
\centering
\caption{LLM Training Loss Across Gradient Strategies}
\label{Loss-LLM}
\begin{tabular}{lccc}
\toprule
\textbf{Optimizer} & \textbf{Standard} & \textbf{Checkpointing} & \textbf{Accumulation} \\
\midrule
Adadelta & 60.54 & 60.51 & \cellcolor{green!25}15.14 \\
Adagrad  & 62.30 & 60.75 & 15.32 \\
Adam     & 118.22 & 73.75 & 22.12 \\
CGD      & 996.59 & 234.31 & 99.39 \\
SGD      & 62.95 & 62.82 & 15.71 \\
\bottomrule
\end{tabular}
\end{table}

% ------------------------ VLM LOSS ------------------------

\begin{table}[H]
\centering
\caption{VLM Training Loss Across Gradient Strategies}
\label{Loss-VLM}
\begin{tabular}{lccc}
\toprule
\textbf{Optimizer} & \textbf{Standard} & \textbf{Checkpointing} & \textbf{Accumulation} \\
\midrule
Adadelta & 0.54 & 0.54 & 0.07 \\
Adagrad  & 0.50 & 0.50 & 0.06 \\
Adam     & 0.66 & 0.66 & \cellcolor{green!25}0.06 \\
SGD      & 0.51 & 0.51 & 0.07 \\
\bottomrule
\end{tabular}
\end{table}

% ================================================================
% GPU UTILIZATION
% ================================================================

\subsection{GPU Utilization}
\begin{figure}[!t]
\centering
\includegraphics[width=\columnwidth]{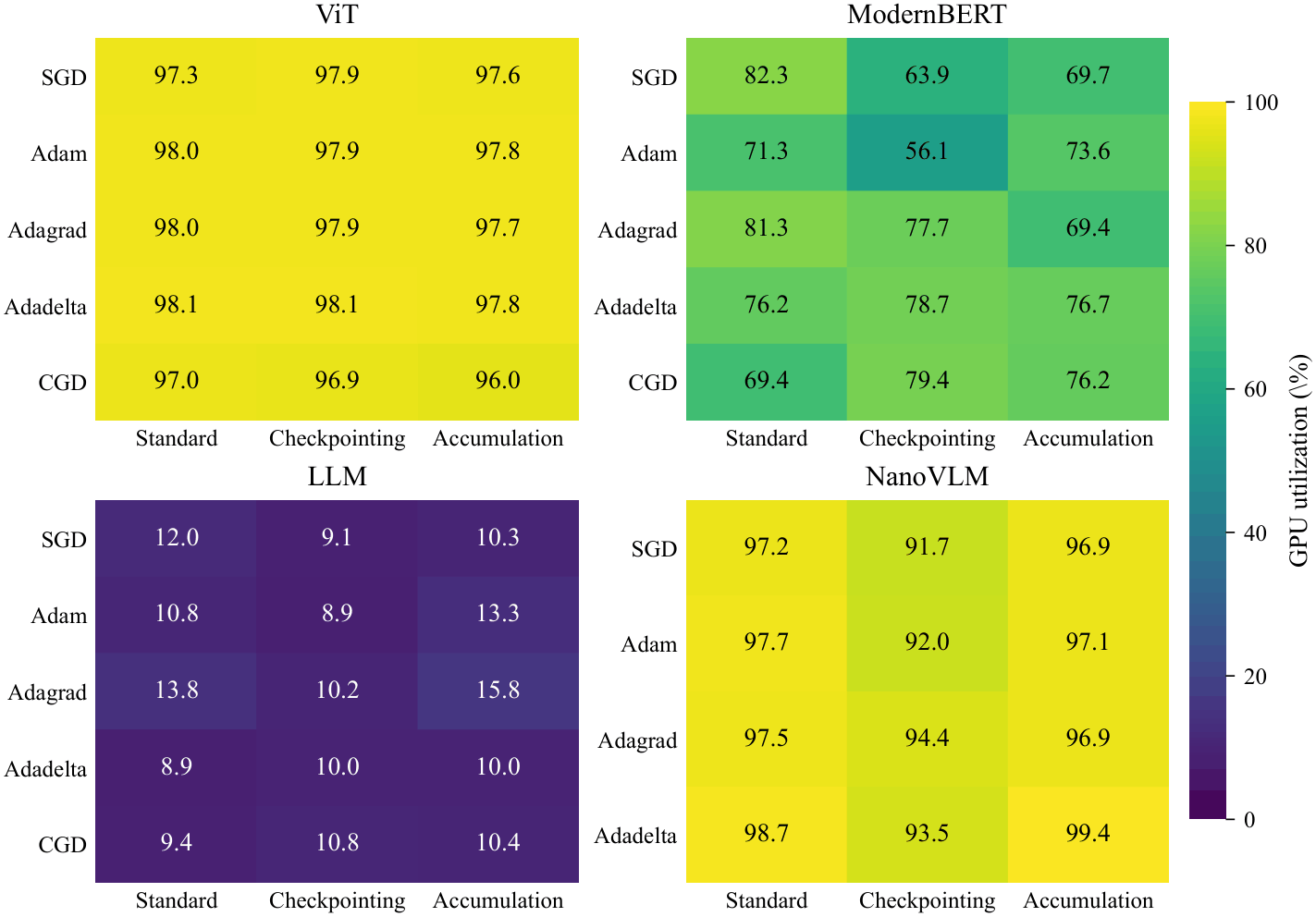}
\caption{Average GPU utilization (\%) across all 25 (pseudo-)epochs for every optimizer--strategy combination. ViT and NanoVLM remain compute-bound at 91--99\% across all configurations; LLM is severely memory-bound at 9--16\%; ModernBERT spans the widest range (56--82\%) and is therefore the most sensitive to configuration choice---most notably, Adam with checkpointing collapses to 56.1\%.}
\label{fig:gpu-heatmap}
\end{figure}

The most significant pattern is the high GPU utilization achieved by both ViT and VLM (96--99\%), compared to ModernBERT (56--82\%) and LLM (8--15\%). This indicates that model architecture and workload characteristics, not optimizer choice, are the primary drivers of GPU efficiency. Figure~\ref{fig:gpu-heatmap} visualizes the full optimizer$\times$strategy grid for each model, immediately revealing the three regimes: compute-bound (ViT, NanoVLM), memory-bound (LLM), and configuration-sensitive (ModernBERT). Let us examine each model.

\textbf{ViT.} All configurations achieve $>$96\% GPU utilization, with most in the 97--98\% range, indicating that ViT's computational pattern (processing image patches through attention) is highly suited to GPU architecture. Adadelta with Standard training achieves 98.14\%, the highest for ViT. Across optimizers, checkpointing and accumulation have negligible impact on utilization ($<$1 percentage point variation), demonstrating that ViT's compute-bound nature absorbs the overhead of both techniques. CGD achieves slightly lower utilization (96--97\%) than other optimizers, likely due to the additional conjugate-direction computation.

\textbf{ModernBERT.} It has a wide utilization range, from 56.08\% (Adam+Checkpointing) to 82.32\% (SGD Standard). This 26-point (56.08-82.32) spread is the largest among all models, suggesting ModernBERT's efficiency is highly sensitive to configuration. SGD Standard is the Efficiency Champion as it achieves 82.32\% —the highest for ModernBERT. This is interesting because SGD is the simplest optimizer—it may have the most predictable, parallelizable computation pattern. Checkpointing Creates Winners and Losers. Indeed, winners are  Adadelta (76.16 $\rightarrow$ 78.68), and  CGD (69.44 $\rightarrow$ 79.40). Also, losers are Adagrad (81.28 $\rightarrow$ 77.72), Adam (71.28 $\rightarrow$ 56.08), SGD (82.32 $\rightarrow$ 63.88). 
Checkpointing's impact is optimizer-dependent. For Adam and SGD, the recomputation steps may create bottlenecks. For CGD, which already has complex second-order computations, the additional recomputation might be better interleaved.
Adam with Checkpointing is the Worst, as 56.08\% is the lowest for ModernBERT—a sharp drop from Adam Standard (71.28\%). This echoes the loss data, where Adam underperformed on ModernBERT. Here we see it's also inefficient.

%Relation to loss, ViT's best loss (0.26) came from Adagrad+Accumulation, which achieves 96.36\% utilization—excellent efficiency alongside top performance. This is an ideal combination.

\textbf{LLM.} Llama\,3.1\,1B exhibits critically low GPU utilization, with all configurations below 16\%. Most fall in the 8--13\% range, meaning the GPU is idle 87--92\% of the time. Despite being small for an LLM, the model appears memory-bound rather than compute-bound: the 1B parameters ($\approx$2\,GB in FP16) plus activations for sequence processing may exceed cache capacity, causing frequent waits on memory transfers. Among all configurations, Adagrad with accumulation achieves the highest utilization at 15.76\%. For most optimizers, checkpointing reduces utilization (Adagrad: 13.76 $\rightarrow$ 10.16; SGD: 12.04 $\rightarrow$ 9.12); only Adadelta sees a tiny improvement (8.92 $\rightarrow$ 9.96). If the model is memory-bound, the additional recomputation from checkpointing may cause cache thrashing, worsening rather than improving compute efficiency.

\textbf{VLM.} NanoVLM achieves high GPU utilization (91--99\%), comparable to ViT. Adadelta with Accumulation reaches 99.44\%, the highest across all VLM configurations. Standard training consistently yields the highest utilization per optimizer (97--99\%), while checkpointing causes a modest but consistent drop of 5--7 percentage points (e.g., SGD: 97.31 $\rightarrow$ 90.90; Adam: 97.60 $\rightarrow$ 92.27). The strong utilization across all configurations reflects NanoVLM's compute-intensive cross-attention mechanism between visual and textual streams, which keeps the GPU well occupied.

% ------------------------ VIT GPU ------------------------

\begin{table}[H]
\centering
\caption{ViT GPU Utilization (\%)}
\label{ViT-GPU}
\begin{tabular}{lccc}
\toprule
\textbf{Optimizer} & Standard & Checkpointing & Accumulation \\
\midrule
Adadelta & 98.14 & 98.12 & 97.81 \\
Adagrad  & 97.97 & 97.94 & 97.69 \\
Adam     & \cellcolor{green!25}98.03 & 97.92 & 97.79 \\
CGD      & 96.97 & 96.85 & 96.01 \\
SGD      & 97.27 & 97.92 & 97.56 \\
\bottomrule
\end{tabular}
\end{table}

% ------------------------ MODERNBERT GPU ------------------------

\begin{table}[H]
\centering
\caption{ModernBERT GPU Utilization (\%)}
\label{ModernBert-GPU}
\begin{tabular}{lccc}
\toprule
\textbf{Optimizer} & Standard & Checkpointing & Accumulation \\
\midrule
Adadelta & 76.16 & 78.68 & 76.72 \\
Adagrad  & 81.28 & 77.72 & 69.40 \\
Adam     & 71.28 & 56.08 & 73.64 \\
CGD      & 69.44 & 79.40 & 76.24 \\
SGD      & \cellcolor{green!25}82.32 & 63.88 & 69.72 \\
\bottomrule
\end{tabular}
\end{table}

% ------------------------ LLM GPU ------------------------

\begin{table}[H]
\centering
\caption{LLM GPU Utilization (\%)}
\label{LLM-GPU}
\begin{tabular}{lccc}
\toprule
\textbf{Optimizer} & Standard & Checkpointing & Accumulation \\
\midrule
Adadelta & 8.92 & 9.96 & 10.00 \\
Adagrad  & 13.76 & 10.16 & \cellcolor{green!25}15.76 \\
Adam     & 10.80 & 8.88 & 13.28 \\
CGD      & 9.36 & 10.80 & 10.44 \\
SGD      & 12.04 & 9.12 & 10.32 \\
\bottomrule
\end{tabular}
\end{table}

% ------------------------ VLM GPU ------------------------

\begin{table}[H]
\centering
\caption{VLM GPU Utilization (\%)}
\label{VLM-GPU}
\begin{tabular}{lccc}
\toprule
\textbf{Optimizer} & Standard & Checkpointing & Accumulation \\
\midrule
Adadelta & 98.69 & 92.28 & \cellcolor{green!25}99.44 \\
Adagrad  & 97.34 & 91.65 & 96.80 \\
Adam     & 97.60 & 92.27 & 96.90 \\
SGD      & 97.31 & 90.90 & 96.88 \\
\bottomrule
\end{tabular}
\end{table}

% ================================================================
% RAM UTILIZATION
% ================================================================

\subsection{System RAM Utilization}
The system RAM usage reveals how different optimization strategies impact host memory (CPU RAM) alongside GPU metrics. This is particularly important for real-world deployments where system memory can become a bottleneck, especially when data loading, preprocessing, or CPU–GPU transfers are involved. In Tables, we represent the RAM usage RAM usage. Overall, it remains stable within each model family, with checkpointing generally increasing RAM due to recomputation overhead. Let's dive into each model's patterns.

\textbf{ViT.} ViT consumes minimal system RAM (8.3--8.5\%), indicating that system memory is not a bottleneck for this architecture. All optimizer and strategy combinations produce nearly identical RAM usage, with SGD consistently the lowest at 8.27--8.28\%. The negligible variation across strategies ($<$0.2 percentage points) suggests that ViT's memory footprint is dominated by model weights rather than optimizer states or activation storage.

\textbf{ModernBERT.} As an efficient encoder, it consumes minimal system memory(18–20\%). It aligns with its design goals and suggests that CPU memory is unlikely to be a bottleneck for this architecture.
Standard Adam training is the most memory-efficient(using 18.64\% RAM, the lowest across all ModernBERT configs). This is surprising given Adam's optimizer states, but perhaps the model's small size and efficient implementation keep the footprint low.
Accumulation drops to 19.00\%, a modest improvement over standard (20.01\%). For other optimizers, accumulation either increases RAM slightly (Adagrad 19.33 $\rightarrow$ 20.04) or leaves it nearly unchanged.
Checkpointing Has Mixed Effects (Adadelta: 20.01 $\rightarrow$ 19.67 (slight decrease), Adam: 18.64 $\rightarrow$ 19.45 (increase), CGD: 19.14 $\rightarrow$ 19.65 (increase), SGD: 19.26 $\rightarrow$ 19.68 (increase)). So, checkpointing does not consistently reduce RAM. Sometimes, it even increases it. This may be because the recomputation logic uses temporary buffers that occupy CPU memory.

\textbf{LLM.} The LLM exhibits the widest variation in system RAM usage (16--38\%), indicating that optimizer and strategy choices have a major impact on host memory. CGD Standard has the lowest RAM footprint at 16.30\%, but this is of little practical value given CGD's very high loss (996.59). Excluding CGD, Adam's standard mode uses 37.77\%---the highest among standard configurations---because its optimizer states (two per parameter) for 1B parameters consume significant memory. Checkpointing sometimes increases RAM substantially (Adadelta: 22.64 $\rightarrow$ 32.75, +45\%; CGD: 16.30 $\rightarrow$ 36.36, +123\%), while for SGD and Adagrad it remains stable ($\pm$1\%). This counterintuitive behavior suggests that for LLMs, the recomputation process may allocate intermediate buffers in system RAM, especially for optimizers with complex state tracking. Accumulation helps optimizers with large state (Adam: 37.77 $\rightarrow$ 32.11; Adadelta: 22.64 $\rightarrow$ 21.18) but can increase RAM for others (Adagrad: 21.89 $\rightarrow$ 24.72; CGD: 16.30 $\rightarrow$ 19.64).

\textbf{NanoVLM.} VLM system memory usage ranges from 3.88\,GB (Adam Standard) to 4.91\,GB (Adadelta Standard). Adam Standard achieves the lowest memory footprint, possibly because its optimizer states are stored entirely on the GPU. Checkpointing and accumulation do not significantly alter system RAM for most optimizers, with variations under 0.2\,GB. The exception is Adam, where standard training uses notably less memory (3.88\,GB) than checkpointing (4.78\,GB) or accumulation (4.24\,GB), suggesting that Adam's reduced optimizer-state pressure on the CPU is partially offset when additional gradient bookkeeping is introduced.

%RAM usage remains stable within each model family, with checkpointing generally increasing RAM due to recomputation overhead.

% ------------------------ VIT RAM ------------------------

\begin{table}[H]
\centering
\caption{ViT RAM Usage (\%)}
\label{ViT-RAM}
\begin{tabular}{lccc}
\toprule
\textbf{Optimizer} & Standard & Checkpointing & Accumulation \\
\midrule
Adadelta & 8.40 & 8.46 & 8.45 \\
Adagrad  & 8.36 & 8.40 & 8.39 \\
Adam     & 8.46 & 8.47 & 8.47 \\
CGD      & 8.49 & 8.50 & 8.50 \\
SGD      & \cellcolor{green!25}8.27 & \cellcolor{green!25}8.27 & \cellcolor{green!25}8.28 \\
\bottomrule
\end{tabular}
\end{table}

% ------------------------ MODERNBERT RAM ------------------------

\begin{table}[H]
\centering
\caption{ModernBERT RAM Usage (\%)}
\label{ModernBERT-RAM}
\begin{tabular}{lccc}
\toprule
\textbf{Optimizer} & Standard & Checkpointing & Accumulation \\
\midrule
Adadelta & 20.01 & 19.67 & \cellcolor{green!25}19.00 \\
Adagrad  & 19.33 & 19.30 & 20.04 \\
Adam     & \cellcolor{green!25}18.64 & 19.45 & 19.50 \\
CGD      & 19.14 & 19.65 & 19.64 \\
SGD      & 19.26 & 19.68 & 19.16 \\
\bottomrule
\end{tabular}
\end{table}

% ------------------------ LLM RAM ------------------------

\begin{table}[H]
\centering
\caption{LLM RAM Usage (\%)}
\label{LLM-RAM}
\begin{tabular}{lccc}
\toprule
\textbf{Optimizer} & Standard & Checkpointing & Accumulation \\
\midrule
Adadelta & 22.64 & 32.75 & 21.18 \\
Adagrad  & 21.89 & 21.40 & 24.72 \\
Adam     & 37.77 & 38.25 & 32.11 \\
CGD      & \cellcolor{green!25}16.30 & 36.36 & 19.64 \\
SGD      & 18.84 & 18.89 & 18.90 \\
\bottomrule
\end{tabular}
\end{table}

% ------------------------ VLM RAM ------------------------

\begin{table}[H]
\centering
\caption{VLM System Memory Usage (GB)}
\label{VLM-RAM}
\begin{tabular}{lccc}
\toprule
\textbf{Optimizer} & Standard & Checkpointing & Accumulation \\
\midrule
Adadelta & 4.91 & 4.75 & 4.74 \\
Adagrad  & 4.84 & 4.90 & 4.89 \\
Adam     & \cellcolor{green!25}3.88 & 4.78 & 4.24 \\
SGD      & 4.81 & 4.83 & 4.82 \\
\bottomrule
\end{tabular}
\end{table}

\subsection{Training Time}
% ================================================================
% TRAINING TIME PER EPOCH
% ================================================================
Across all four models, two clear trends emerge. First, checkpointing generally increases training time (by up to 60\% depending on model and optimizer), though the penalty is negligible for most ViT configurations. This is expected since recomputing activations adds computational work. Second, accumulation often reduces training time compared to standard training, sometimes substantially (e.g., ViT: 42s $\rightarrow$ 32s). Fewer optimizer updates can offset the overhead of gradient accumulation. Let us examine each model in detail.

\textbf{ViT.} SGD is the fastest optimizer, achieving the lowest training times in both Standard (31.77s) and Accumulation (31.74s) modes. Accumulation reduces time by roughly 23\% for the adaptive optimizers (Adadelta: 43.95 $\rightarrow$ 34.22, 22\% faster; Adagrad: 42.10 $\rightarrow$ 32.35, 23\% faster; Adam: 42.74 $\rightarrow$ 33.00, 23\% faster). Because these adaptive optimizers have per-parameter update rules that consume CPU/GPU time, accumulation reduces the frequency of these updates, lowering overhead while processing the same number of samples.
Checkpointing has minimal impact on time for most optimizers. As shown in Table~\ref{ViT-Time}, for Adadelta, Adagrad, Adam, and CGD, checkpointing adds $<$0.1s (negligible), while for SGD it increases time by 30\% (31.77 $\rightarrow$ 41.43s). For ViT, the recomputation cost of checkpointing is largely hidden by parallelism---except for SGD. SGD's simplicity may make it harder to overlap recomputation with other operations, or the optimizer step itself is so fast that recomputation becomes the new bottleneck. Finally, CGD is surprisingly fast. Despite being a second-order method, CGD is only slightly slower than Adam in Standard mode (43.47 vs 42.74s), suggesting the conjugate gradient iterations are efficiently implemented.

\textbf{LLM.} Checkpointing imposes significant overhead, increasing time by 45--58\% across all optimizers (Adadelta: 22.70 $\rightarrow$ 35.81, +58\%; SGD: 22.51 $\rightarrow$ 32.64, +45\%). Because the LLM is already memory-bound (GPU utilization 8--15\%), recomputation adds serial work that cannot be hidden by parallelism, directly extending runtime. Accumulation provides modest gains: most optimizers see a 2--5\% speedup (e.g., Adadelta: 22.70 $\rightarrow$ 22.02). SGD accumulation is fastest overall at 21.90\,s. The reduction in optimizer update frequency helps, but the memory-bound nature of the model limits gains compared to compute-bound models like ViT. As shown in Table~\ref{LLM-Time}, CGD Standard at 28.40\,s is 21\% slower than CGD Accumulation (23.38\,s), and CGD Checkpointing at 39.50\,s is the slowest configuration overall---consistent with its very high loss (996.59) and low utilization.
%—second-order methods are a poor fit for autoregressive LLMs.
 
\textbf{ModernBERT.} Accumulation delivers major speedups based on Table \ref{ModernBERT-Time}, (Adadelta: 25.45 $\rightarrow$ 20.25 (20\% faster), Adagrad: 21.65 $\rightarrow$ 19.17 (11\% faster), Adam: 23.18 $\rightarrow$ 19.47 (16\% faster), CGD: 21.42 $\rightarrow$ 19.17 (11\% faster). Our interpretation is that  ModernBERT, as an efficient encoder, likely has lightweight forward/backward passes, making optimizer overhead a larger fraction of total time. %Reducing update frequency via accumulation pays off handsomely.
SGD is at the highest speed, and SGD Standard with  18.69s is the fastest standard. SGD accumulation with 18.60s is the fastest overall.
Checkpointing is uniformly slower: all optimizers see time increase by 5--25\% with checkpointing. As shown in Table~\ref{ModernBERT-Time}, the largest increase is for SGD (18.69 $\rightarrow$ 26.96, +44\%). SGD's simplicity may make recomputation overhead more visible. This matches the ModernBERT loss data, where checkpointing caused severe degradation (loss increased 7--16$\times$).

\textbf{VLM.} NanoVLM's total training time (Table~\ref{VLM-Time}) reveals a clear split between optimizers. SGD (27.10\,min), Adam (27.63\,min), and Adagrad (27.64\,min) cluster tightly, while Adadelta is substantially slower at 37.85\,min---a 40\% overhead. Checkpointing uniformly increases total time to 41\,min regardless of optimizer, indicating that recomputation dominates over optimizer-specific overhead. Accumulation has negligible effect on training time for most optimizers, with SGD+Accumulation achieving the fastest total time at 26.76\,min.

%The following tables report the average time required to complete one training epoch for different optimizers and gradient strategies. Checkpointing consistently increases epoch time due to recomputation overhead, while accumulation often achieves the fastest training.

% ------------------------ VIT TIME ------------------------

\begin{table}[H]
\centering
\caption{ViT Training Time per Epoch (seconds)}
\label{ViT-Time}
\begin{tabular}{lccc}
\toprule
\textbf{Optimizer} & Standard & Checkpointing & Accumulation \\
\midrule
Adadelta & 43.95 & 43.97 & 34.22 \\
Adagrad  & 42.10 & 42.15 & 32.35 \\
Adam     & 42.74 & 42.76 & 33.00 \\
CGD      & 43.47 & 43.44 & 33.80 \\
SGD      & \cellcolor{green!25}31.77 & 41.43 & \cellcolor{green!25}31.74 \\
\bottomrule
\end{tabular}
\end{table}

% ------------------------ LLM TIME ------------------------

\begin{table}[H]
\centering
\caption{LLM Training Time per Epoch (seconds)}
\label{LLM-Time}
\begin{tabular}{lccc}
\toprule
\textbf{Optimizer} & Standard & Checkpointing & Accumulation \\
\midrule
Adadelta & 22.70 & 35.81 & 22.02 \\
Adagrad  & 22.96 & 33.89 & 22.56 \\
Adam     & 22.84 & 33.57 & 22.38 \\
CGD      & 28.40 & 39.50 & 23.38 \\
SGD      & 22.51 & 32.64 & \cellcolor{green!25}21.90 \\
\bottomrule
\end{tabular}
\end{table}

% ------------------------ MODERNBERT TIME ------------------------

\begin{table}[H]
\centering
\caption{ModernBERT Training Time per Epoch (seconds)}
\label{ModernBERT-Time}
\begin{tabular}{lccc}
\toprule
\textbf{Optimizer} & Standard & Checkpointing & Accumulation \\
\midrule
Adadelta & 25.45 & 26.92 & 20.25 \\
Adagrad  & 21.65 & 26.85 & 19.17 \\
Adam     & 23.18 & 26.99 & 19.47 \\
CGD      & 21.42 & 26.87 & 19.17 \\
SGD      & 18.69 & 26.96 & \cellcolor{green!25}18.60 \\
\bottomrule
\end{tabular}
\end{table}

% ------------------------ VLM TIME ------------------------

\begin{table}[H]
\centering
\caption{VLM Total Training Time (minutes, single epoch of 18{,}623 steps)}
\label{VLM-Time}
\begin{tabular}{lccc}
\toprule
\textbf{Optimizer} & Standard & Checkpointing & Accumulation \\
\midrule
Adadelta & 37.85 & 41.40 & 38.34 \\
Adagrad  & 27.64 & 41.28 & \cellcolor{green!25}26.91 \\
Adam     & 27.63 & 41.42 & 27.00 \\
SGD      & 27.10 & 41.02 & 26.76 \\
\bottomrule
\end{tabular}
\end{table}

\section{Discussion and Key Findings}

\begin{figure*}[!t]
\centering
\includegraphics[width=\linewidth]{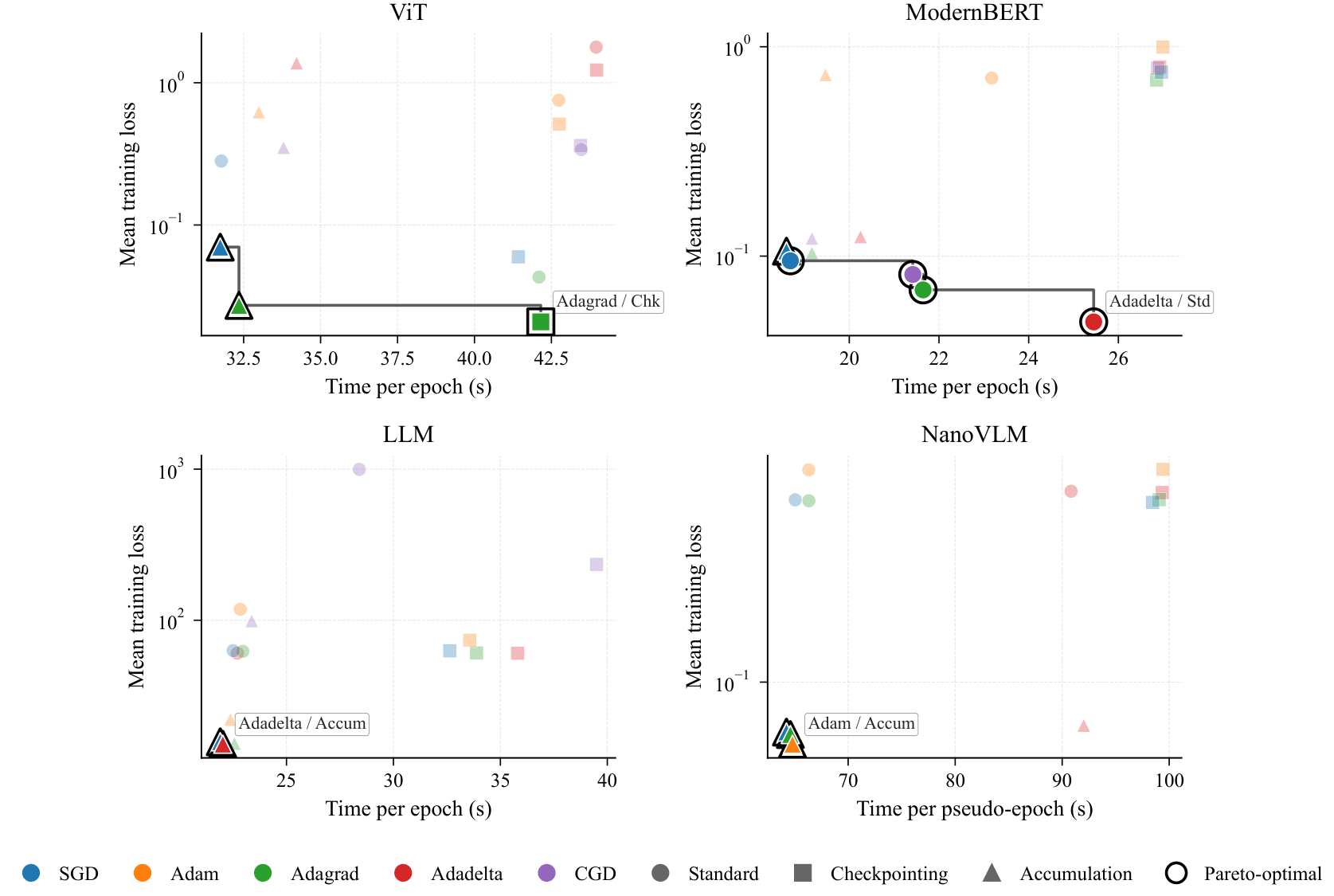}
\caption{Final training loss (mean of last five epochs, log-scale) versus average time per (pseudo-)epoch, for every (optimizer, strategy) configuration. Color encodes optimizer; marker shape encodes strategy. Pareto-optimal points are highlighted with a black ring and connected by a stair-step line. Gradient accumulation is on the Pareto frontier for all four architectures.}
\label{fig:pareto}
\end{figure*}

To frame the configuration-selection problem from a practitioner's perspective, Figure~\ref{fig:pareto} plots final loss against per-(pseudo-)epoch wall-clock time for every (optimizer, strategy) combination, with the Pareto frontier highlighted. The shape of the frontier differs sharply across architectures: ViT exhibits a wide trade-off curve (faster configurations cost up to $5\times$ in loss), ModernBERT collapses to a tight cluster of accumulation/standard configurations near the optimum (with checkpointing pushed far from the front), LLM shows a clean accumulation-dominated frontier with all standard and checkpointing points strictly dominated, and NanoVLM presents a binary outcome---accumulation-mode points form the entire frontier. We unpack these patterns below.

The recommendation flow chart introduced in Figure~\ref{fig:best-combos} crystallises these trade-offs into per-metric guidance. Several patterns emerge at a glance. SGD is the dominant choice for training time and memory (seven of sixteen best paths, including the fastest configuration on every model). Adadelta wins the loss tournament on ModernBERT and Llama\,3.1, Adagrad on ViT, and Adam on NanoVLM. Gradient accumulation lies on half of all best paths and is the consistently preferred strategy for training time. Conjugate Gradient Descent, in contrast, is never the best choice for any (model, metric) combination---which is why it is absent from the optimizer column---reinforcing the conclusion that CGD is dominated everywhere in our benchmark.

\textbf{Gradient accumulation consistently improves or maintains model performance across all architectures.}
Across LLMs, ViT, VLM, and ModernBERT, accumulation produced the lowest or near-lowest loss values. In VLM, accumulation reduces average loss by roughly an order of magnitude (e.g., Adam: 0.66$\rightarrow$0.06). For the LLM, loss dropped from above 60 in standard mode to around 15 with accumulation. ViT benefits more modestly, as optimizers like SGD and Adagrad already achieve low losses in standard mode. ModernBERT, being inherently stable, showed nearly identical loss across modes, but accumulation never degraded performance. This improvement occurs because accumulation effectively increases the batch size without increasing memory usage, producing smoother gradients and more stable optimization.

\textbf{Adaptive optimizers show mixed results depending on architecture.}
Adam consumes the highest system RAM on LLMs (37.77\% standard), due to storing additional moment and variance tensors for 1B parameters. However, Adam is not always the best performer: it is 14$\times$ worse than Adadelta on ModernBERT, 2$\times$ worse on LLMs, and 19$\times$ worse than Adagrad on ViT (standard mode). Adadelta, despite also being adaptive, achieves the best or near-best loss on ModernBERT, LLMs, and is competitive on VLM and ViT, suggesting its moving-window approach to gradient statistics provides more robust behavior than Adam's fixed decay across architectures.

\textbf{CGD performance is highly model-dependent.}
CGD achieved strong performance on ViT (loss 0.34 standard, better than Adam's 0.75) and ModernBERT (loss 0.08 standard, near the best), but diverged sharply on LLMs with a standard loss of 996.59---more than 15$\times$ worse than any other optimizer. Even with accumulation, CGD's LLM loss (99.39) is 5--7$\times$ worse than Adam, Adadelta, SGD, or Adagrad (15.14--22.12). Interestingly, checkpointing produces the largest CGD improvement of any optimizer on LLMs, reducing loss by 76\% (996.59 $\rightarrow$ 234.31), though still leaving it unusable. This instability on LLMs stems from CGD's sensitivity to curvature and batch variance---small batch sizes or high-curvature regions cause oscillation or divergence. CGD consistently uses very little GPU memory ($\approx$2\,GB across all LLM configurations), making it attractive for compact vision and encoder models with smoother loss surfaces.

\textbf{Checkpointing shows architecture-dependent effects on loss but consistently increases runtime.}
Across experiments, checkpointing substantially slowed training (by 20--60\% depending on model) due to recomputation overhead. Its effect on loss varies sharply by model: for ViT, checkpointing consistently reduces loss across all optimizers (e.g., SGD: 0.28$\rightarrow$0.06; Adagrad: 0.04$\rightarrow$0.02), suggesting a regularization benefit. For VLM, the effect is neutral (ratios near 1.00). For ModernBERT, checkpointing causes severe loss degradation (7--16$\times$ worse). For LLMs, checkpointing is generally neutral for SGD/Adadelta/Adagrad but notably benefits Adam (38\% loss reduction) and CGD (76\% reduction). GPU utilization also dips during LLM checkpointing runs, indicating increased CPU--GPU synchronization in a memory-bound workload. Checkpointing is thus beneficial for compute-bound vision models, situationally useful for LLMs with specific optimizers, but should be avoided for ModernBERT-style encoders.

\textbf{Gradient accumulation acts as an implicit regularizer.}
Across all models, accumulation not only smoothed gradients but also reduced overfitting tendencies, especially in ViT and NanoVLM. Loss curves showed less variance across epochs, even without explicit regularization such as weight decay or dropout adjustments. This aligns with theoretical findings that larger batch sizes approximate more stable gradient directions.

\textbf{CGD shows promise in vision and encoder models.}
CGD achieved competitive loss on ViT (0.34 standard, comparable to Adam) and ModernBERT (0.08 standard, near the best). Its training time was close to Adam (43.47 vs 42.74s on ViT), indicating efficient implementation. However, CGD diverged sharply on LLMs (loss near 1000). This indicates that CGD benefits models with smoother loss surfaces, and may be worth exploring for distilled or compact vision/encoder models.

\textbf{Overall, gradient accumulation emerges as the most reliable and efficient method.}  
It reduces loss, stabilizes training, avoids memory spikes, and offers improvements across almost all settings. Adaptive optimizers provide fast convergence but require careful memory budgeting. First-order methods are memory-friendly and perform well when combined with accumulation. CGD is a niche and unstable for large models. Checkpointing, while helpful for memory-restricted hardware, does not meaningfully improve loss.
%%%%%%%%%%%%%%%%%%%%%%%%%%%%%%%%%%%%%%%%%%%%%%%%%%%%%%%%%%%%%%
\section{Conclusion}
%%%%%%%%%%%%%%%%%%%%%%%%%%%%%%%%%%%%%%%%%%%%%%%%%%%%%%%%%%%%%%
On-device training of large models remains constrained by limited memory and computation. Across ViT, ModernBERT, Llama,3.1,1B, and NanoVLM, evaluated with five optimizers and three gradient--memory strategies, gradient accumulation was the most broadly effective technique. It reduced LLM loss from over 60 to approximately 15 and VLM loss by roughly an order of magnitude while maintaining performance on ViT and ModernBERT. Adadelta was the most consistent optimizer across architectures, whereas Adam performed up to 14$\times$ worse than Adadelta on ModernBERT, 2$\times$ worse on LLMs, and 19$\times$ worse than Adagrad on ViT. CGD remained competitive on ViT (0.34) and ModernBERT (0.08) but diverged on the LLM (996.59), demonstrating its sensitivity to batch variance and curvature in autoregressive models. Checkpointing was strongly architecture-dependent: it improved ViT and selected LLM configurations, had little effect on the VLM, severely degraded ModernBERT, and consistently increased runtime.

For memory-constrained edge devices, SGD or Adagrad with gradient accumulation provides stable training with low overhead, while Adadelta with accumulation is preferable when rapid and consistent convergence is required. These results also show that decoder-only LLMs are more optimizer-sensitive than vision and encoder models and that memory strategies should reflect the underlying bottleneck: LLM GPU utilization remained at 8--15%, indicating memory-bound execution, whereas ViT and VLM reached 96--99%, indicating compute-bound workloads. Overall, gradient accumulation acts as an implicit regularizer that stabilizes training, Adadelta is the most reliable adaptive optimizer, CGD should be limited to vision and encoder architectures, and checkpointing should be applied selectively. Future work will evaluate quantized and 8-bit training, dynamic optimizer switching, hybrid memory--compute techniques, larger models, and real-world edge hardware to develop more deployable on-device fine-tuning pipelines.

\bibliographystyle{plain}
\bibliography{refs}

\end{document}